\documentclass[default,iicol]{sn-jnl}
\usepackage{adjustbox}
\usepackage{ulem}

\jyear{2021}%

\theoremstyle{thmstyleone}%
\theoremstyle{thmstyletwo}%

\theoremstyle{thmstylethree}%

\begin{document}

\title[Article Title]{Table Structure Recognition with Conditional Attention}

\author[1]{\fnm{Bin} \sur{Xiao}}\email{bxiao103@uottawa.ca}
\author[1]{\fnm{Murat} \sur{Simsek}}\email{murat.simsek@uottawa.ca}
\author[1]{\fnm{Burak} \sur{Kantarci}}\email{burak.kantarci@uottawa.ca}
\author[2]{\fnm{Ala Abu} \sur{Alkheir}}\email{ala\_abualkheir@lytica.com}

\affil[1]{\orgname{School of Electrical Engineering and Computer Science}, \orgaddress{\street{University of Ottawa}, \city{Ottawa}, \postcode{K1N 6N5}, \state{ON}, \country{Canada}}}

\affil[2]{\orgname{Lytica}, \orgaddress{\street{308 Legget Dr}, \city{Ottawa}, \postcode{K2K 1Y6}, \state{ON}, \country{Canada}}}

\abstract{
Tabular data in digital documents is widely used to express compact and important information for readers. However, it is challenging to parse tables from unstructured digital documents, such as PDFs and images, into machine-readable format because of the complexity of table structures and the missing of meta-information. Table Structure Recognition (TSR) problem aims to recognize the structure of a table and transform the unstructured tables into a structured and machine-readable format so that the tabular data can be further analysed by the down-stream tasks, such as semantic modeling and information retrieval. In this study, we hypothesize that a complicated table structure can be represented by a graph whose vertices and edges represent the cells and association between cells, respectively. Then we define the table structure recognition problem as a cell association classification problem and propose a conditional attention network (CATT-Net). The experimental results demonstrate the superiority of our proposed method over the state-of-the-art methods on various datasets. Besides, we investigate whether the alignment of a cell bounding box or a text-focused approach has more impact on the model performance. Due to the lack of public dataset annotations based on these two approaches, we further annotate the ICDAR2013 dataset providing both types of bounding boxes, which can be a new benchmark dataset for evaluating the methods in this field. Experimental results show that the alignment of a cell bounding box can help improve the Micro-averaged F1 score from 0.915 to 0.963, and the Macro-average F1 score from 0.787 to 0.923. 
}

\keywords{Table Structure Recognition, Conditional Attention Mechanism, Tabular Data Understanding.}

\maketitle
\section{Introduction}
\label{sec:intro}
Tabular data is a popular format to express compact and important information for readers in different types of digital documents, including web pages, PDFs, Word processors, and document images~\cite{qiao2021lgpma}. With the development of NLP techniques, table semantic understanding tasks, such as Table-based Question Answering tasks~\cite{vakulenko2017tableqa,talmor2020multimodalqa} and table-based entity Linking tasks ~\cite{bhagavatula2015tabel,Luo2018CrossLingualEL}, have drawn considerable attention for the research community. However, these recent studies only consider relational and structured table data which are only a small portion of tables in all types of digital documents ~\cite{wang2021tuta}. Therefore, it is vital to transform unstructured tabular data into structured and machine-readable format to bridge the gap between the state-of-the-art NLP algorithms and the data from the real-world. Table Structure Recognition (TSR) problem aims at recognizing the structure of a table and transform the unstructured tables into a structured and machine-readable format. Since almost all types of digital documents can be easily converted into images, we focus on the TSR problem on document images in this study. It is a challenging task because document images can only provide visual clues without any meta-data~\cite{tensmeyer2019deep}. Moreover, a complicated table structure can have many variations, such as spanning cells, multi-row or multi-column headers, partially bordered or even none-bordered cells, making it challenging. 

\begin{figure*}[htp]
\begin{center}
  \includegraphics[width=0.7\textwidth]{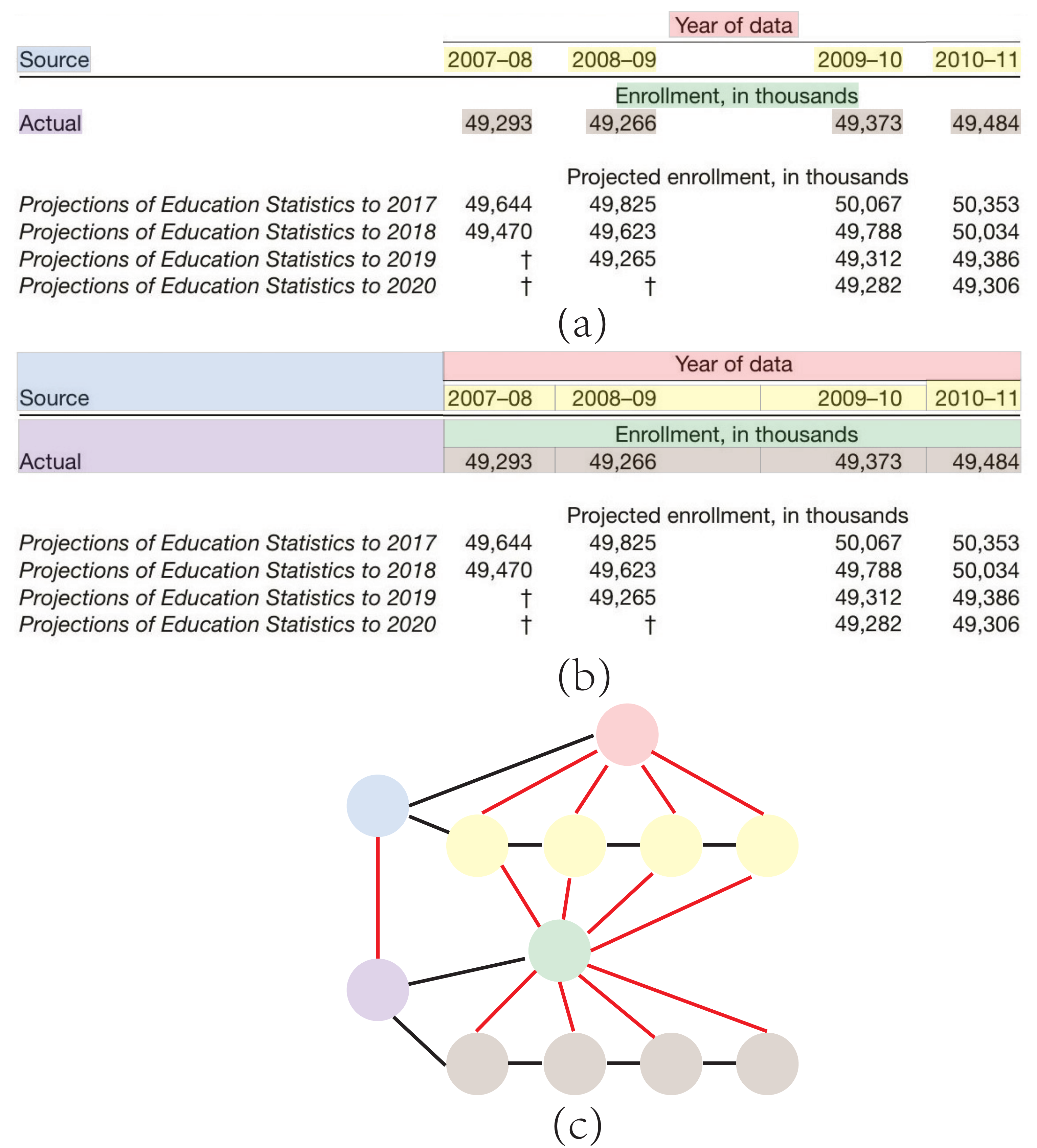}
  \caption{Table examples with different types of bounding boxes. In Figure(a), cells' bounding boxes are not aligned and only focus on the text areas, while in Figure(b) bounding boxes are aligned. Figure(c) is an example of using a graph to represent part of the complicated table structure of Figure(a) and Figure(b). Red lines represent vertical connections and black lines mean horizontal connections.}
  \label{fig:table_examples}
\end{center}
\end{figure*}

With the development of deep learning, there have been many deep learning based approaches trying to solve the table structure recognition problem and achieved promising results. Some of these approaches~\cite{tensmeyer2019deep,gilani2017table} leverage top-down methods and identify columns and rows using semantic segmentation and object detection methods. Meanwhile, some studies~\cite{li2020gfte,chi2019complicated} employ bottom-up approaches and represent the table structures with graph models. More specifically, bottom-up approaches assume that each table and its cells' bounding boxes are known, then the cells can be represented by the vertexes in a graph, and the edges in the graph can represent the association between two cells. Thus, to build the table structure, we only need to categorize the edges into three types of associations: "vertical connection", "horizontal connection", and "no connection". Although these studies have achieved promising results dealing with the data from a single data source, the models of these studies often have limited generalization abilities because the document images can come from different domains with different resolution and table styles. Position features, such as the cell bounding boxes and cell coordinates, are effective when the training set and the testing set are from a single source and share an identical coordinate system.  But, at the same time, they also limit the generalization ability of the model when we set the experiment in a cross-domain setting. Besides, the performance of bottom-up models heavily relies on the cell bounding boxes. Figure~\ref{fig:table_examples} shows two types of cell bounding boxes, in which the first type only covers the text area and is not aligned; in contrast, the second type is aligned based on the table structure. Most public datasets use the aligned bounding boxes, but it is not easy to extract the aligned bounding box when the table does not have explicit borders, such as the table in the Figure~\ref{fig:table_examples}.
Since there have already been many successful table detection and table cell detection solutions, we assume that the bounding boxes of tables and those of cells are given in this study. Therefore, the contributions of this work are three-fold: 1) Formulation of the table structure recognition task as a cell association prediction problem so to develop a bottom-up model, namely Conditional Attention Network (CATT-Net). The experimental results show that CATT-Net can outperform the state-of-the-art methods without using a large amount of training data. Moreover, CATT-Net can also perform well in the cross domain setting because it only utilizes visual features avoiding the adverse effects of position features when the different domains are using different coordinate systems or have different image resolutions.
2) A conditional attention module by using the two cell images in a cell pair as the condition of both channel attention and spatial attention, leading the model to learn more distinctive features so to build the association of the two cells via the attention mechanism. 3) Performance evaluation of the two types of bounding box annotations in the context of the table structure recognition problem. This is also complemented by a new annotation of the ICDAR2013 dataset with text-focused bounding boxes so to provide both types of bounding boxes.

The rest of the paper is structured as follows. First, we discuss the related work including table detection and table structure in section~\ref{sec:related work}. Second, we elaborate the problem definition and the proposed method in section~\ref{sec:proposed_method}. Then we present the experiments and the discussion in section~\ref{sec:experiments}. At last, in section~\ref{sec:conclusion}, we draw our conclusions and discuss the future directions of this topic.
\section{Related Work}
\label{sec:related work}
Table detection, cell detection and table structure recognition problems have been widely discussed in recent years. For PDF documents, heuristics-based approaches are popular and achieved promising performance. Typically, heuristics methods~\cite{yildiz2005pdf2table, shigarov2016configurable} need to define various rules and use the meta-data of the documents, meaning that this type of approach cannot process document images. Besides, the generalization ability of heuristics-based methods are often limited because of variations of table structures. With the development of deep learning, deep learning based approaches are becoming the dominant solutions, therefore, we only focus on deep learning based approaches in this section.

\subsection{Table detection}
Table detection and cell detection tasks can be treated as the pre-process step towards the table structure recognition, and have been widely discussed in many studies. Typically, table detection and table structure recognition are often defined as object detection problems and semantic segmentation problems. Therefore, popular object detection solutions, including Faster-RCNN~\cite{ren2016faster}, Mask-RCNN~\cite{He2020MaskR}, and semantic segmentation solutions, such as FCN~\cite{long2015fully}, are widely used in related studies.
DeepDeSRT~\cite{schreiber2017deepdesrt} is a deep learning based approach that aims at detecting tables and recognizing table structures. The authors use transfer learning and carefully fine-tune a Faster R-CNN model and a FCN semantic segmentation model for table detection and table recognition. Similarly, Traquair et al.~\cite{traquair2019deep} employ Faster-RCNN and RetinaNet~\cite{lin2017focal} to detect tables and use transfer learning to further improve the model performance and reduce training time. Agarwal et al.~\cite{agarwal2021cdec} also propose a deep learning network for table detection termed CDeC-Net using a multi-stage extension of Mask R-CNN. CascadeTabNet~\cite{prasad2020cascadetabnet} is another deep learning network aiming at detecting tables and recognizing the structural body cells at the same time. It uses a cascade network architecture and heavily rely on the data augmentation and transfer learning techniques. TableNet~\cite{paliwal2019tablenet} tries to solve the table detection and structure recognition problem with a single model based on semantic segmentation. There are also other studies using similar object detection and semantic segmentation methods, but to sum up, most of these studies are using a top-down approach, designing new network architectures and using transfer learning methods to achieve better performance. For top-down approaches, table detection algorithms are often combined with table structure recognition tasks such as column and row detection. Meanwhile, for bottom-up approaches, table detection is a necessary step before table cell detection, and the complex table structures are built on the associations of table cells. Hence, in this study we use table detection as a step before cell detection because the proposed method is a bottom-up approach.

\begin{figure*}[htp]
\begin{center}
  \includegraphics[width=\textwidth]{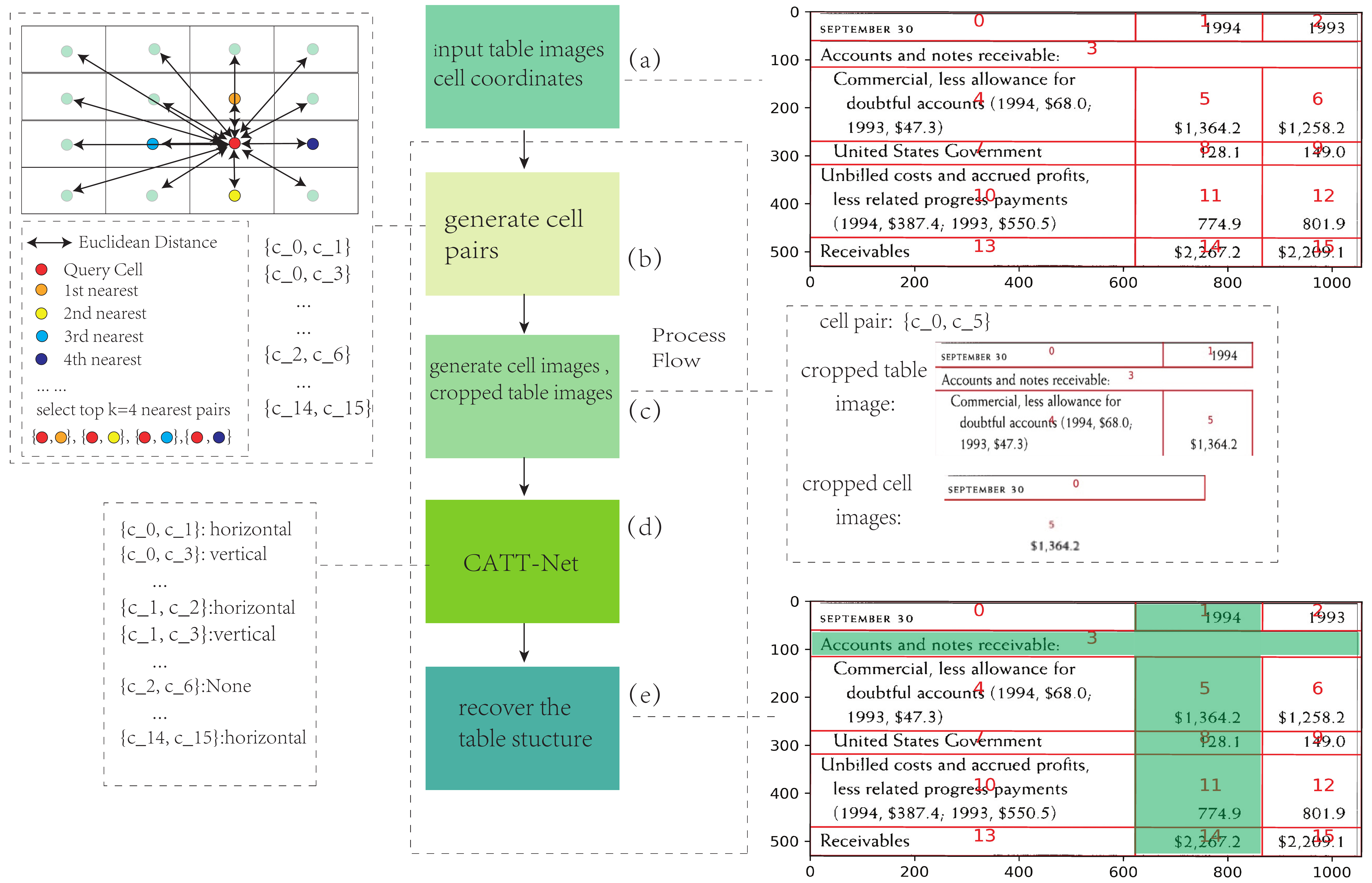}
  \caption{A minimalist illustration of the process and data flow of the proposed method. Figure (a) is the original table and each cell in the table is represented by a sequence number. Figure (b) represents the step of generating cell pairs, which will be discussed in section~\ref{sec:conditional_attention_network}. In this step, we use a k-nearest algorithm to reduce time complexity of generating cell pairs. Figure (c) is the step where cropped table images are generated based on the cell pairs. Figure (d) is the proposed proposed CATT-Net, which takes the cropped table image and the cell images as the inputs, and can output the associations of cell pairs. Figure (e) shows an example of recovering the table structure by given a query cell, then outputting the cells in the same column and row with the query cell.}
  \label{fig:flowchart}
\end{center}
\end{figure*}
\subsection{Table structure recognition}
Table structure recognition tasks are often combined with the table detection tasks by using top-down approaches, such as TableNet~\cite{paliwal2019tablenet}. Kara et al.~\cite{kara2020holistic} propose to use MASK-RCNN to detect the columns and rows in the tables directly so as to obtain the table structure. Qiao et al.~\cite{qiao2021lgpma} propose a Local and Global Pyramid Mask Alignment framework by applying the soft pyramid mask learning mechanism to local and global feature maps. The authors argue that the aligned bounding boxes of text region are efficient features to describe the cells' range and their relation, but are difficult to be predicted. On the other hand, we can also use bottom-up approaches to solve the table structure recognition problem. Typically, bottom-up approaches rely on the cells’ bounding box and often uses the graph to represent tables. More specifically, bottom-up approaches usually assume that the cells' bounding boxes are known and the cells are the basic units of tables. Then the cells and cells' relations can be represented by the edges and vertexes of the graph model, respectively. For example, Chi et al. ~\cite{li2020gfte} propose a graph-based approach named GraphTSR using a graph attention mechanism,  but they only considered the cell position features. Similarly, Li et al.~\cite{li2020gfte} propose a graph convolution based model named GFTE, but this approach integrates more features, including image feature, position feature and textual feature~\cite{chi2019complicated}. Our proposed method also uses a graph structure to represent the complex table structure but unlike the existing bottom-up models, it uses bounding boxes to generate cropped images, and proposes a conditional attention mechanism so that the model can extend the generalization ability of features from bounding boxes and learning more discriminate visual features.

\section{Proposed Method}
\label{sec:proposed_method}
In this section, we present the problem description alongside the proposed method. A minimalist flow of the proposed method is presented in Figure~\ref{fig:flowchart} where the steps of table processing is presented at the high level. The following subsections elaborate on the details of the proposed method.

\subsection{Problem Definition}
\label{sec:problem definition}
Following the definition in the study~\cite{chi2019complicated}, a table with a complicated structure can be represented by a graph $\mathbf{G}$ consisting of a vertex set $\mathbf{V}$ and an edge set $\mathbf{E}$, namely $\mathbf{G}=<\mathbf{V}, \mathbf{E}>$. The set of vertices $\mathbf{V}$ can be used to represent the cells in a table, and the set of edges $\mathbf{E}$ can be used to represent the associations of cells. We define three types of relations between two cells: "vertical connection", "horizontal connection", and "no connection". Figure~\ref{fig:table_examples} shows an example of representing a table with a graph where the red lines mean vertical connection; the black lines mean horizontal connection. Formally, given a training set containing $N$ tables $\mathbf{T}=\{t_i\}_{i=1}^{i=N}$, in each table, the association $r_{\{j,k\}}$ of the cell pair $\{c_j, c_k\}$, $r_{\{j,k\}} \in \{0, 1, 2\}$, where $c_j$ is the $jth$ cell and $c_k$ is the $kth$ cell in an identical table, and $0, 1, 2$ represent no connection, vertical connection, and horizontal connection, respectively. For every cell pair $\{c_j, c_k\}$ composed of the $j$th cell and the $k$th cell in an identical table, the model is to determine the probability $P_\theta (r_{\{j,k\}}=\hat{r}_{\{j,k\}}\rvert(c_j, c_k)), \hat{r}_{\{j,k\}}\in\{0, 1, 2\}$, where $\theta$ represents the trainable parameters of the model. We briefly summarize the notations presented in this paper in Table ~\ref{table:notation_table}

\subsection{Conditional Attention Network for TSR}
\label{sec:conditional_attention_network}
\begin{figure*}[htp]
\begin{center}
  \includegraphics[width=\textwidth]{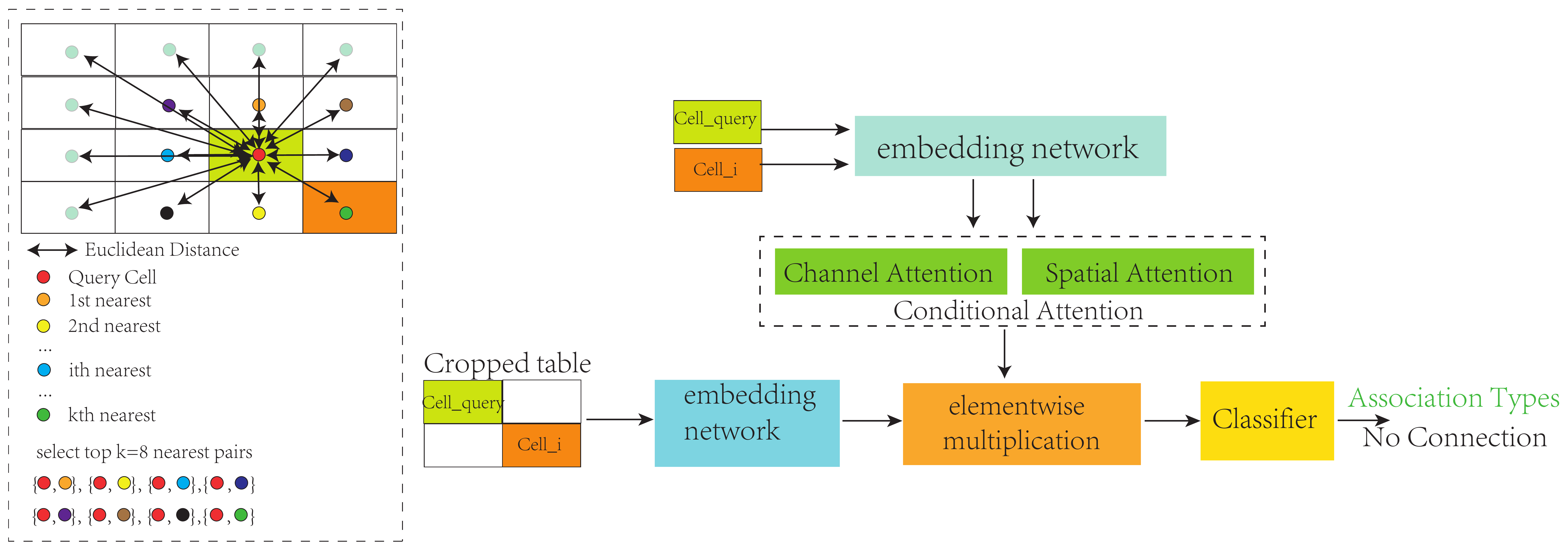}
  \caption{The architecture of the proposed method. The proposed model takes a cropped image and two cell images as inputs, and outputs the association types of the two cells. In this figure, two cells are not associated, hence, the model's output is "No Connection".
  }
  \label{fig:overall_architecture}
\end{center}
\end{figure*}

As discussed in the previous section, we can use a graph to represent a complicated table structure and define the table structure recognition problem classifying the relation of two cells from an identical table. However, if we simply iterate all combinations of the cells in a table, the time complexity would be $O(M^2)$, where $M$ is the number of cells in the table, meaning that it can be very time-consuming when the $M$ is large. Therefore, we use a k-nearest approach by only considering the top $K$ nearest cell to form cell pairs. We use Euclidean distance as the metric to measure the distance of two cells, and KD-Tree can quickly implement this approach. Notably, $K$ is a hyper parameter here, and we set $K$ to 20 in our method.
\begin{table*}[ht!]
\caption{A summary of notations used in this paper.}
\centering
\begin{tabular}{ c c}
\hline
\label{table:notation_table}
 $\mathbf{G}$ & A graph model consists of a vertex set and a edge set\\
 $\mathbf{E}$ & The edge set in a graph model \\
 $\mathbf{V}$ & The vertex set in a graph model \\
 $\mathbf{T}$ & A table set \\
 $\mathbf{C}$ & The cell set of a table\\
 $\{x_i^1, x_i^2, y_i^1, y_i^2\}$ & The coordinate of the $i$th cell in a table\\
 $c_i$ & The $i$th cell in a table\\
 $t_i$ & The $i$th table in a table set $\mathbf{T}$\\
 $r_{\{j,k\}}$ & The relation(association) of the $j$th cell and the $k$th in a table\\
 $e_i$ & The feature embedding of the $i$th cell in a table\\
 \hline
\end{tabular}
\end{table*}
The overall architecture of our method is shown in Figure~\ref{fig:overall_architecture}. The proposed model takes three inputs, including two cell images and a cropped table image, and the image features generated by the two cell images act as the inputs of the proposed Conditional Attention Module; then the output attention of the the proposed Conditional Attention Module is multiplied with the features generated by the cropped table as the final feature fed into the classifier. Instead of simply feeding the whole table images into the embedding network as many other studies, we firstly crop the images of cells in each cell pair, and also crop the part of the image containing the corresponding cells. Suppose the $jth$ cell and the $kth$ cell in the table $t_i$ form the cell pair $\{c_j, c_k\}$, and the coordinates of $c_j$ and $c_k$ are $(x_j^1, x_j^2, y_j^1, y_j^2)$ and $(x_k^1, x_k^2, y_k^1, y_k^2)$ respectively, then the coordinates of the cropped table image should be $(min(x_j^1, x_k^1), max(x_j^1, x_k^1), min(y_j^1, y_k^1), max(y_j^2, y_k^2))$. Thus, we can extract the cropped cell images and table images based on the coordinates information.

We use a simple ConvNet4~\cite{vinyals2016matching} as the embedding network to extract features from images, as shown in Figure~\ref{fig:overall_architecture}. It is worth to note that there are two separate embedding networks that do not share parameters in the proposed method. One is used to extract the features of cell images and the other one is designed to  extract the features of cropped table images. Since this work uses a CNN to extract the visual features and fully connected layers as the classifier, the model requires fixing the size of the input images. For cell and table images, their appearance and visual features can be easily influenced by resizing the shape leading to an adverse effect to the model performance, especially when the height and width ratios are different. Therefore, to keep the ratio of height and width, we use a padding method instead of simply resizing the images in the pre-processing step. More specifically, assuming the input size is $(h_{in}, w_{in})$ and the target size is $(h_{out}, w_{out})$, where $h$ means the height of the image, $w$ denotes the width of the image, if the target size is larger than the input size, meaning that $h_{out} > h_{in}$ and $w_{out}> w_{in}$, then we simply pad zeros to the input image, otherwise a resizing ration can be determined by $ratio = max(h_{out} / h_{in}, w_{out} / w_{in})$. We use this resizing $ratio$ to resize the input images firstly, then use zero paddings to make the input images have the target size. 

The proposed Conditional Attention Module consists of two parts: Conditional Channel Attention and Conditional Spatial Attention. Both Channel Attention and Spatial Attention take the two cells' features as the input. Assuming that the feature embedding pair generated by the cell pair $\{c_l, c_m\}$ is $\{e_l, e_m\}$, $e_l \in \mathbb{R}^{c_{in} \times h_{in} \times w_{in}}$, $e_m \in \mathbb{R}^{c_{in} \times h_{in} \times w_{in} }$, and the feature embedding generated by the corresponding table image is $e_{\{l,m\}} \in \mathbb{R}^{c_{out} \times h_{out} \times w_{out} }$, where $c, h, w$ are the number of channels, the height and the width, respectively. The Conditional Channel Attention can be define as the sum of a trainable function $f_\phi$ with two different inputs, as shown in Equation~\ref{eq:channel_attention}, where the output channel attention $a_{ca} \in \mathbb{R}^{c_{out} \times 1 \times 1 }$.

\begin{equation}
\label{eq:channel_attention}
\begin{aligned}
a_{ca} = {f_\phi}(e_l) + {f_\phi}(e_m), l \in \{1,2,\ldots,K\}, \\ m \in \{1,2,\ldots,K\}, l \neq m 
\end{aligned}
\end{equation}

\begin{figure*}[htp]
\begin{center}
  \includegraphics[width=\textwidth]{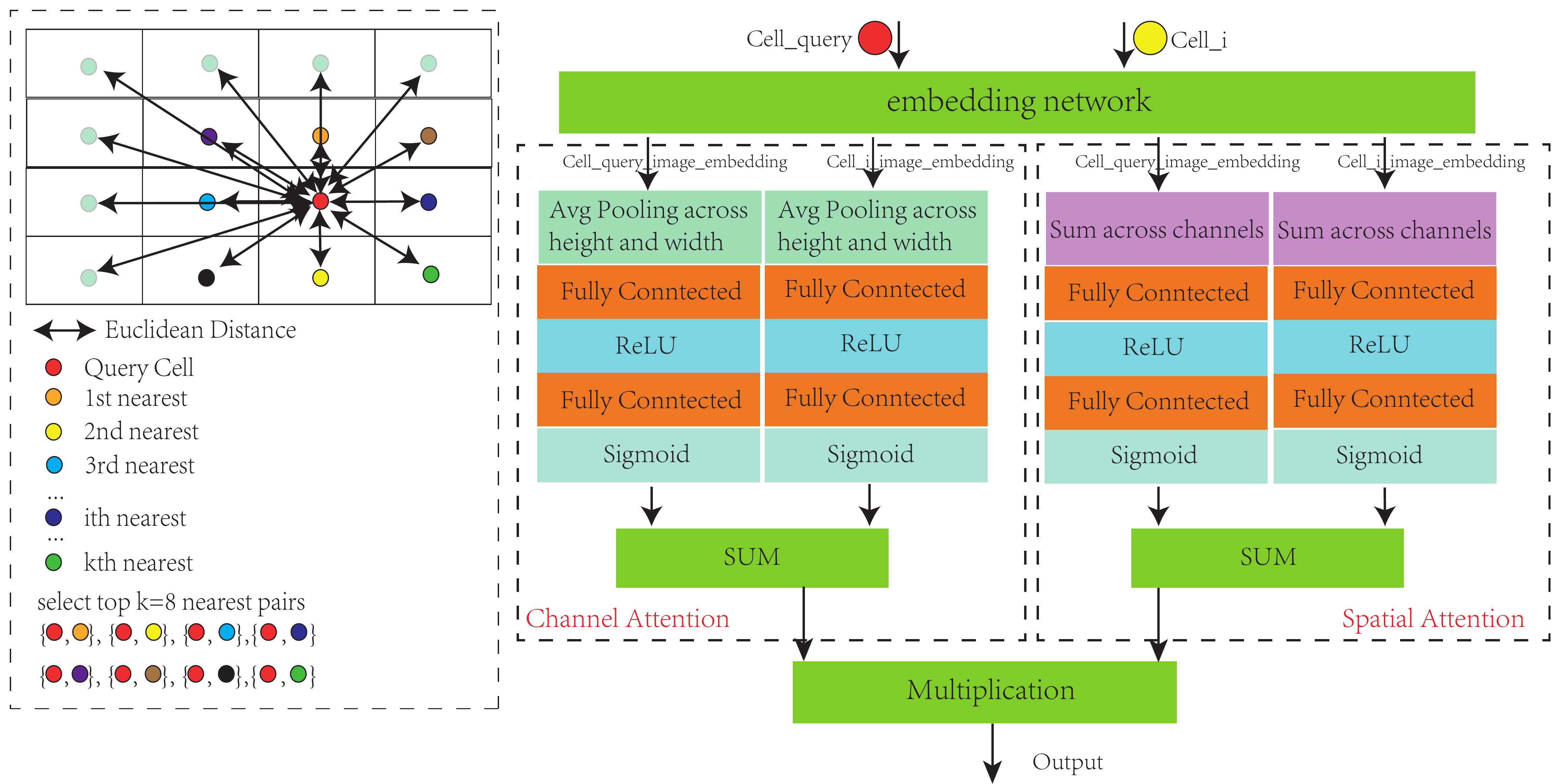}
  \caption{The proposed conditional attention module. "Cell$_i$ feature"  denotes an output of the embedding network}
  \label{fig:conditional_attention}
\end{center}
\end{figure*}

Similarly, the Conditional Spatial Attention also can be define as the sum of a trainable function $g_\tau$ with embedding pair $\{e_l, e_m\}$ as inputs, as shown in Equation~\ref{eq:spatial_attention}, where the output $a_{sp} \in \mathbb{R}^{1 \times h_{out} \times w_{out}}$.
\begin{equation}
\label{eq:spatial_attention}
\begin{aligned}
a_{sp} = {g_\tau}(e_l) + {g_\tau}(e_m), l \in \{1,2,\ldots,K\}, \\ m \in \{1,2,\ldots,K\}, l \neq m
\end{aligned}
\end{equation}
We implement both trainable function $f_\phi$  and $g_\tau$ by a small network consists of two fully connected layers, as shown in Figure~\ref{fig:conditional_attention}, but the output fully connected layers of them have different number of neurons. In the Conditional Channel Attention, the last fully connected layer contains $c_{out}$ units, whereas in the Conditional Spatial Attention, the last fully connected layer has $h_{out} * w_{out}$ units. The output of both parts are expand to the shape $c_{out} * h_{out} * w_{out}$ and multiplied together as the final output of this module. At last, we use cross entropy as the loss function of the proposed model, and the model can be trained in a end-to-end manner.

\begin{figure}[htp]
\begin{center}
  \includegraphics[width=\columnwidth]{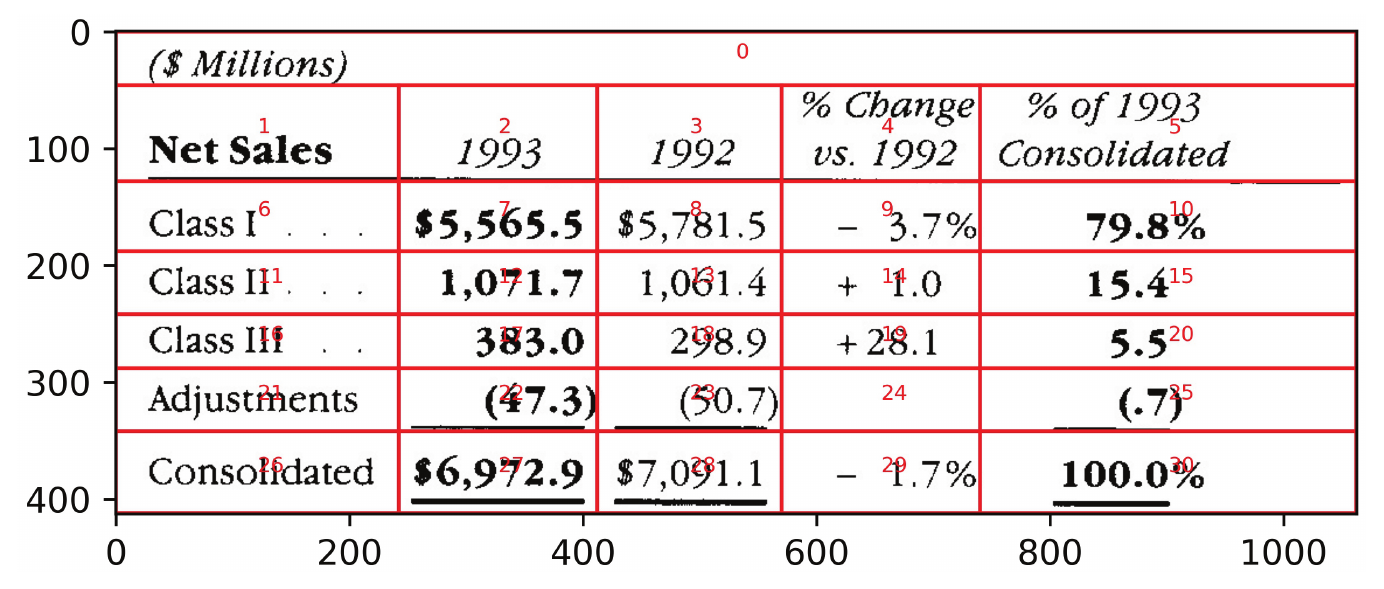}
  \caption{A sample table containing an empty cell from the UNLV dataset}
  \label{fig:empty_cell_example}
\end{center}
\end{figure}
\subsection{Recover the table structure from graph}
In this section, we discuss how to recover the complicated table structure from the output of our method. As stated earlier, we assume that the coordinates of cells in the table are known. In practise, there have been some studies~\cite{jiang2021tabcellnet, kara2020holistic} that can help us obtain the coordinates of the cells with promising performance. Besides, we can obtain the associations of the cell pairs in the table through the proposed method. Therefore, the table structure can be recovered on the basis of the cell coordinates and associations between the cells. We use Algorithm~\ref{alg:structure_recover_algo} to illustrate how to find the cells in the first row. The output of the Algorithm~\ref{alg:structure_recover_algo} is a cell set $\mathbf{C}={\{c_1, c_2, ..., c_n}\}$ containing $n$ cells that are in the first row. These cells can be sorted based on their coordinates to recover their order in the original table.

\begin{algorithm}
\label{alg:structure_recover_algo}
\caption{Finding the cells Cell Set $\mathbf{C}$ in the first row}
\hspace*{\algorithmicindent} \textbf{Input:} The associations of table cells \\
\hspace*{\algorithmicindent} \textbf{Output:} A set of cells in the first row
\begin{algorithmic}[1]
\State Initialize an empty queue $Q$

\State Initialize an empty set $S$

\State Find the cell with minimum $y^1$, push the cell into $Q$

\While{Q.size() $>$ 0}
\State Pop the top cell $c_t$ from $Q$

\State Add $c_t$ to the $S$

\State Find $c_t$ horizontal connected cell list $l_c$

\State k = len($l_c$) - 1

\While{k $>=$ 0}
  \If{$l_c[k]$ not in $S$}
  \State Push $l_c[k]$ into $Q$
  \EndIf
  \State k = k -1
  \EndWhile
\EndWhile
\State $C$ = $S$
\end{algorithmic}
\end{algorithm}

It is worth mentioning that we follow multiple policies to deal with the empty cells as types of bounding boxes vary. For aligned bounding boxes, empty cells are often also annotated because they can reflect the visual and logical structure of the table, whereas, for text-focused bounding boxes, empty cells should be ignored. Therefore, when text-focused bounding boxes are used, it is still considered that two cells are connected when there are empty cells between them. For example, Figure~\ref{fig:empty_cell_example} shows a table from the UNLV dataset, which contains an empty cell. The red lines are the aligned bounding boxes and the red numbers are the \textit{id}s of cells. One can find that the $24th$ cell ($c_{24}$) is an empty cell but it is still annotated when aligned bounding boxes are used. However, it should be discarded when text-focused bounding boxes are used. It is worth to note that in the case of text-focused bounding boxes, $c_{23}$ and $c_{25}$ would be defined as horizontally connected, $c_{19}$ and $c_{29}$ would be defined as vertically connected. Thus, the table structure can still be recovered despite the empty cells when text-focused bounding boxes are used.
\begin{table*}[ht!]
\caption{Macro-averaged results on ICDAR2013, SciTSR, and SciTSR-COMP. CATT-Net represents our proposed method.}
\centering
\begin{tabular}{  c  c |c | c || c | c | c || c | c | c }
\hline
\label{table:macro_averaged_experimental_results}
& \multicolumn{3}{c}{ICDAR2013} & \multicolumn{3}{c}{SciTSR} & \multicolumn{3}{c}{SciTSR-COMP}\\ 
\hline
\hline
Method & Precision & Recall & F1 & Precision & Recall & F1 & Precision & Recall & F1\\
\hline
 \texttt{Tabby} & $0.789 $ & $0.845 $ & $0.816 $ & $0.914 $ & $0.910 $ & $0.912 $ & $0.869 $ & $0.841 $ & $0.855 $\\
 \texttt{GraphTSR} & $0.819 $ & $0.855 $ & $0.837 $ & $0.936 $ & $0.931 $ & $0.934 $ & $0.943 $ & $0.925 $ & $0.934 $\\
 \texttt{DeepDeSRT} & $0.573 $ & $0.564 $ & $0.568 $ & $0.898 $ & $0.897 $ & $0.897 $ & $0.811 $ & $0.813 $ & $0.812 $\\
 \texttt{CATT-Net} & $\mathbf{0.941} $ & $\mathbf{0.907} $ & $\mathbf{0.923} $ & $\mathbf{0.956} $ & $\mathbf{0.965} $ & $\mathbf{0.961} $ & $\mathbf{0.966} $ & $\mathbf{0.961}$ & $\mathbf{0.964} $\\
 \hline
\end{tabular}
\end{table*}

\begin{table*}[ht!]
\caption{Micro-averaged results on ICDAR2013, SciTSR, and SciTSR-COMP. CATT-Net represents our proposed method.} 
\centering
\begin{tabular}{  c  c |c | c || c | c | c || c | c | c }
\hline
\label{table:micro_averaged_experimental_results}
& \multicolumn{3}{c}{ICDAR2013} & \multicolumn{3}{c}{SciTSR} & \multicolumn{3}{c}{SciTSR-COMP}\\  
\hline
\hline
Method & Precision & Recall & F1 & Precision & Recall & F1 & Precision & Recall & F1\\
\hline
 \texttt{Tabby} & $0.846 $ & $0.862 $ & $0.854 $ & $0.926 $ & $0.920 $ & $0.921 $ & $0.892 $ & $0.872 $ & $0.882 $\\
 \texttt{GraphTSR} & $0.885 $ & $0.860 $ & $0.872 $ & $0.959 $ & $0.948 $ & $0.953 $ & $0.964 $ & $0.945 $ & $0.955 $\\
 \texttt{TabStrucNet} & $0.915 $ & $0.897 $ & $0.906 $ & $0.927 $ & $0.913 $ & $0.920 $ & $0.909 $ & $0.882 $ & $0.895$ \\
 \texttt{Split+Heur} & $0.938 $ & $0.922 $ & $0.930 $ & $-$ & $-$ & $-$ & $-$ & $-$ & $-$ \\ 
 \texttt{DeepDeSRT} & $0.632 $ & $0.617 $ & $0.615 $ & $0.906 $ & $0.887 $ & $0.890 $ & $0.863 $ & $0.831 $ & $0.846 $ \\
 \texttt{CATT-Net} & $\mathbf{0.963} $ & $\mathbf{0.963} $ & $\mathbf{0.963} $ & $\mathbf{0.981} $ & $\mathbf{0.981} $ & $\mathbf{0.981}  $ & $\mathbf{0.982} $ & $\mathbf{0.982}$ & $\mathbf{0.982} $ \\
 \hline
\end{tabular}
\end{table*}
\section{Experiments and Analysis}
\label{sec:experiments}
\subsection{Datasets}
We use four public datasets, including ICDAR2013~\cite{gobel2013icdar}, UNLV~\cite{unlvdataset}, SciTSR~\cite{chi2019complicated},  and SciTSR-COMP~\cite{chi2019complicated} to evaluate the proposed method.
ICDAR2013 dataset consists of 158 tables, meanwhile, UNLV, SciTSR and SciTSR-COMP dataset contain 558, 15000, and 716 tables respectively. Besides, we provide another type of text-focused bounding boxes for ICDAR2013, which can be used to compare its effectiveness to the proposed method with the aligned bounding boxes.
\begin{table}[ht!]
\caption{Statistics of datasets. UNLV dataset is used as the training set of ICDAR2013 dataset.}
\centering
\begin{tabular}{  | c |c | c | c |c | }
\hline
\label{table:statistics_of_datasets}
   & training & validation & testing & total\\ 
\hline
 \texttt{ICDAR2013} & $ - $ & $ 126 $ & $ 32 $ & $ 158 $ \\
 \texttt{UNLV} & $ - $ & $ - $ & $ - $ & $ 558 $ \\
 \texttt{SciTSR} & $ 9000 $ & $ 3000 $ & $ 3000 $ & $ 15000 $\\
 \texttt{SciTSR-COMP} & $ 430 $ & $ 143 $ & $ 143 $ & $ 716 $\\ 
 \hline
\end{tabular}
\end{table}
\subsection{Experiment settings and results}
\label{sec:experiment_results}
We follow the evaluation protocol and experiment settings presented by~\cite{chi2019complicated} to conduct experiments on the datasets mentioned above. More specifically, in the experiments, precision, recall, and F1 score are used as the evaluation metrics, and the UNLV dataset is used as the training set for the experiments on the ICDAR2013 dataset because the ICDAR2013 dataset does not provide a training set. The ICDAR2013 dataset is split into a validation set containing $20\%$ of its tables and a test set containing $80\%$ of its tables. For other datasets, each of them is split into a training set, a validation set, and a test set based on a $60:20:20$ split. The statistics of the datasets are briefly summarized in Table~\ref{table:statistics_of_datasets}.  In each experiment, we use the validation set to fine-tune the hyper parameters and determine the best model. For the embedding network, we use a simple ConvNet4~\cite{vinyals2016matching} which consists of 4 convolution layers with 64 filters in each convolution layer. Each convolution layer is followed by a batch normalization layer~\cite{ioffe2015batch}, a ReLU layer~\cite{nair2010rectified}, and a max pooling layer.The optimizer used in our implementation is AdamW~\cite{Loshchilov2019DecoupledWD} with an initial learning rate 0.001. All the images are resized and padded to 84 * 84 using the padding method discussed in the previous section. Micro-averaged and macro-averaged scores are used to calculated the evaluation metrics, which are defined in Equations~\ref{eq:macro_f1}~\ref{eq:micro_f1}~\ref{eq:micro_precision}~\ref{eq:micro_recall}. It is worth to note that $\{F1_i, Precision_i, Recall_i\}$ in the Equation~\ref{eq:macro_f1} represent the $\{F1, Precision, Recall\}$ scores of the $i$th class. Thus, macro-averaging treats all classes equally regardless of their sample size whereas micro F1, micro precision and micro recall in the Equations~\ref{eq:micro_f1}~\ref{eq:micro_precision}~\ref{eq:micro_recall} are performed by firstly calculating the sum of all true positives and false positives over all the classes followed by the calculation of the precision and recall for the sums~\cite{micro_average}.

\begin{equation}
\label{eq:macro_f1}
\begin{aligned}
\text{Macro} \{\text{F1}, \text{Precision}, \text{Recall}\} \\ = \frac{1}{N}\sum_{i=0}^{N}\{\text{F1}_i,\text{Precision}_i, \text{Recall}_i\}
\end{aligned}
\end{equation}

\begin{equation}
\label{eq:micro_f1}
\begin{aligned}
\text{Micro F1} = 2*\frac{\text{MicroPrecision} * \text{MicroRecall}}{\text{MicroPrecision} + \text{MicroRecall}}
\end{aligned}
\end{equation}

\begin{equation}
\label{eq:micro_precision}
\begin{aligned}
\text{MicroPrecision} = \frac{\text{True positive}}{\text{True positive} + \text{False positive}}
\end{aligned}
\end{equation}

\begin{equation}
\label{eq:micro_recall}
\begin{aligned}
\text{MicroRecall} = \frac{\text{True positive}}{\text{True positive} + \text{False negative}}
\end{aligned}
\end{equation}

The experimental results are shown in Table~\ref{table:macro_averaged_experimental_results} and Table~\ref{table:micro_averaged_experimental_results} which are macro averaged and micro averaged scores. Tabby~\cite{shigarov2018tabbypdf} is a rule-based approach that is designed to process PDF documents, meaning that it cannot deal with document images. GraphTSR~\cite{chi2019complicated} is a graph-based approach heavily relying on the position features, including the cell coordinates, cell center, cell size, and table size, but without using any visual features. DeepDeSRT~\cite{schreiber2017deepdesrt}  uses semantic segmentation to segment the rows and columns in a table, meaning that it only relies on the visual features. TableStrucNet~\cite{raja2020table} considers two experiment settings in their work. The first setting only uses visual features, whereas the second setting also leverages other features in their model, including bounding boxes and other meta information. Split+Heur~\cite{tensmeyer2019deep} contains two deep learning models: one is termed Split that can predict basic table grid patterns, anther is Merge that can decide whether grid elements should be merged as span cells. Moreover, Split+Heur also uses a heuristic method to define various rules as the post-processing step to improve the results. Notably, the experimental results of Tabby, GraphTSR, DeepDeSRT on the ICDAR2013, SciTSR and SciTSR-COMP dataset come from the study~\cite{chi2019complicated}, and the results of TableStrucNet and Split+Heur are from study~\cite{raja2020table}. We only list the results of TableStrucNet's first setting for a fair comparison because our model does not use other meta-information directly. Experimental results show that the proposed method can outperform those benchmark models.
\subsection{Analysis and Discussion}
\label{sec:analysis_and_discussion}
\subsubsection{Ablation study}
We conduct an ablation study to analysis effectiveness of the components in our method. First, to prove whether the proposed model can benefit from the conditional attention module, we compare the proposed model with its counterpart model without using the conditional attention module. Second, as we consider the position features implicitly in the pre-processing step of our approach, we also compare the model only using cell position features. The experimental results are shown in Table ~\ref{table:ablation_study}. Notably, CATT-Net-POS means the model that only uses position features,  CATT-Net-NOATT represents the model without using the conditional attention module, and CATT-Net is the proposed method, and we use UNLV dataset as the training set, 20\% of ICDAR2013 dataset as the validation set, and the rest of ICDAR2013 as the testing set. The experimental results show that visual features are more effective than position features when the experiments are conducted in this cross-domain setting. In contrast, when the training set and testing set are from a single source, position features can also be very effective and achieve prosming results. As shown in Table~\ref{table:macro_averaged_experimental_results} and Table~\ref{table:micro_averaged_experimental_results}, GraphTSR~\cite{chi2019complicated} is a model that only taking position features as inputs and achieves better performance than some visual features based solutions proving the effectiveness of position features, even though its performance is not better than our proposed method. Besides, our proposed Conditional Attention Module can help improve the performance of the model by a large margin.

\begin{table*}[ht!]
\caption{Experimental results of using different components in the proposed method.}
\centering
\begin{tabular}{  c| c |c | c || c |c | c }
\hline
\label{table:ablation_study}
& \multicolumn{3}{c}{Macro averaged} & \multicolumn{3}{c}{Micro averaged}\\ 
\hline
  Embedding Network & Precision & Recall & F1 & Precision & Recall & F1\\
\hline
 \texttt{CATT-Net-POS} & $0.751 $ & $0.914 $ & $0.764 $ & $0.799 $ & $0.799 $ & $0.799 $\\
 \texttt{CATT-Net-NOATT} & $0.882 $ & $0.878 $ & $0.878 $ & $0.943 $ & $0.943 $ & $0.943 $\\
 \texttt{CATT-Net} & $\mathbf{0.941} $ & $\mathbf{0.907} $ & $\mathbf{0.923} $ & $\mathbf{0.963} $ & $\mathbf{0.963}$ & $\mathbf{0.963} $\\ 
 \hline
\end{tabular}
\end{table*}

\subsubsection{Embedding Network}
As discussed in the previous section, our proposed method heavily rely on the visual features and position features. To explore the impact of different visual features generated by different embedding network, we conduct experiments on the UNLV and ICDAR2013 datasets. More specifically, we use UNLV dataset as the training set and 20\% of ICDAR2013 as the validation set and the rest of ICDAR2013 as the testing set, and three embedding networks, including ConvNet4, ResNet12 and ResNet34 are compared. We use the validation set to fine-tune the hyper parameters and determine the best model in the experiments, and the experimental results are shown in Table ~\ref{table:embedding_network_comparsions}. Increasing the depth of the embedding network can help improve the performance to some extent, but the model cannot always benefit from increasing the depth.

\begin{table*}[ht!]
\caption{Experimental results of using different embedding networks.}
\centering
\begin{tabular}{  c| c |c | c || c |c | c }
\hline
\label{table:embedding_network_comparsions}
  & \multicolumn{3}{c}{Macro averaged} & \multicolumn{3}{c}{Micro averaged}\\ 
  \hline
  Embedding Network & Precision & Recall & F1 & Precision & Recall & F1\\
\hline
 \texttt{ConvNet4} & $0.941 $ & $0.907 $ & $0.923 $ & $0.963 $ & $0.963$ & $0.963 $\\
 \texttt{ResNet12} & $0.942 $ & $\mathbf{0.951} $ & $\mathbf{0.946} $ & $\mathbf{0.973} $ & $\mathbf{0.973} $ & $\mathbf{0.973} $\\
 \texttt{ResNet34} & $\mathbf{0.947} $ & $0.937 $ & $0.942 $ & $0.972 $ & $0.972 $ & $0.972 $\\ 
 \hline
\end{tabular}
\end{table*}

\subsubsection{The impact of training set size}
Deep learning based approaches often require a large dataset making it challenging to collect sufficient labeled data and increase the training overhead. In this section, we compare the model performance by varying the number of training tables from the UNLV dataset while other experiment settings remain as explained earlier. The experimental results are presented in 
Table~\ref{table:number_of_training_comparsions}. As seen in the table, increasing the number of training samples can help improve the performance when the number of training samples is significantly small. However, using larger number of training tables does not guarantee improvement in the model performance. Besides, our proposed method can achieve promising results using a relatively small datatset because of its bottom-up nature that does not rely on object detection or semantic segmentation. In contrast, solutions based on object detection and segmentation often need large training sets, such as SciTSR ~\cite{chi2019complicated} which contains 12000 training samples and 3000 testing samples.  

\begin{table*}[ht!]
\caption{Experimental results of using different number of training tables.}
\centering
\begin{tabular}{  c| c |c | c || c |c | c }
\hline
\label{table:number_of_training_comparsions}
  & \multicolumn{3}{c}{Macro averaged} & \multicolumn{3}{c}{Micro averaged}\\ 
  \hline
  Number of training tables & Precision & Recall & F1 & Precision & Recall & F1\\
\hline
 \texttt{300} & $0.934 $ & $\mathbf{0.938} $ & $\mathbf{0.936} $ & $\mathbf{0.970} $ & $\mathbf{0.970} $ & $\mathbf{0.970} $\\
 \texttt{250} & $0.922 $ & $0.935 $ & $0.928 $ & $0.968 $ & $0.968 $ & $0.968 $\\
 \texttt{200} & $\mathbf{0.943} $ & $0.927 $ & $0.935 $ & $\mathbf{0.970} $ & $\mathbf{0.970} $ & $\mathbf{0.970} $\\
 \texttt{150} & $0.935 $ & $0.918 $ & $0.926 $ & $0.967 $ & $0.967 $ & $0.967 $\\
 \texttt{100} & $0.926 $ & $0.923 $ & $0.923 $ & $0.965 $ & $0.965 $ & $0.965 $\\
 \texttt{50} & $0.913 $ & $0.897 $ & $0.903 $ & $0.954 $ & $0.954$ & $0.954 $\\
 \hline
\end{tabular}
\end{table*}

\subsubsection{Comparison of two types of bounding boxes}
Since in the pre-processing step, we use the bounding boxes to generate the input images, including the cell images and the cropped table images. In this section, we compare the impact of two types of bounding boxes to the proposed method. As shown in Figure~\ref{fig:table_examples}, the first type of bounding box is the text-focused bounding boxes that only cover text area in each cell, whereas the second type is the aligned bounding boxes that can reflect the cell relation and table structure. To the best of our knowledge, there is no public dataset providing these two types of bounding boxes simultaneously, so we further annotate the ICDAR2013 dataset with the text-focused bounding boxes. Aligned bounding boxes are more effective than text-focused bound boxes because aligned bounding boxes can reflect the visual structure of the table. However, aligned bounding boxes are hard to obtain for the object detection and semantic segmentation methods, especially when the table does not have borders or only have partial borders. In contrast, object detection and semantic segmentation methods can infer the text-focused bounding boxes much easier, but these bounding boxes can lose the table structural information. We conduct experiments on the ICDAR2013 dataset keeping the same experiment settings with the experiments in the previous section to compare the impact of the two types of bounding boxes to the proposed method, and the experimental results are shown in Table ~\ref{table:bounding_boxes_comparsions}. We can find that the model using the aligned bounding boxes can outperform its counterpart model using text-focused bounding boxes significantly.
\begin{table*}[ht!]
\caption{Experimental results of using different types of bounding boxes.}
\centering
\begin{tabular}{  c| c |c | c || c |c | c }
\hline
\label{table:bounding_boxes_comparsions}
  & \multicolumn{3}{c}{Macro averaged} & \multicolumn{3}{c}{Micro averaged}\\ 
  \hline
  Type of bounding boxes & Precision & Recall & F1 & Precision & Recall & F1\\
\hline
 \texttt{Aligned} &  $\mathbf{0.941} $ & $\mathbf{0.907} $ & $\mathbf{0.923} $ & $\mathbf{0.963} $ & $\mathbf{0.963}$ & $\mathbf{0.963} $\\
 \texttt{Text-focused} & $0.848 $ & $0.759 $ & $0.787 $ & $0.914 $ & $0.915 $ & $0.915 $\\
 \hline
\end{tabular}
\end{table*}

\begin{figure*}[htp]
\begin{center}
  \includegraphics[width=\textwidth]{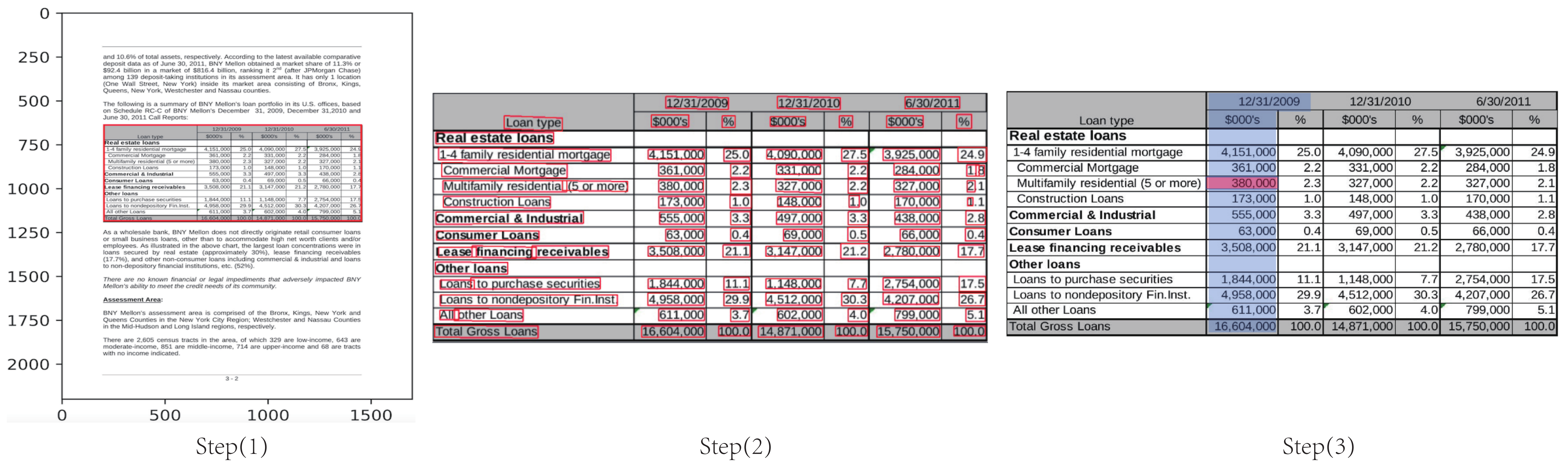}
  \caption{Three steps of the full pipeline.}
  \label{fig:demonstration_of_proposed_method}
\end{center}
\end{figure*}

\begin{figure}[htp]
\begin{center}
  \includegraphics[width=\columnwidth]{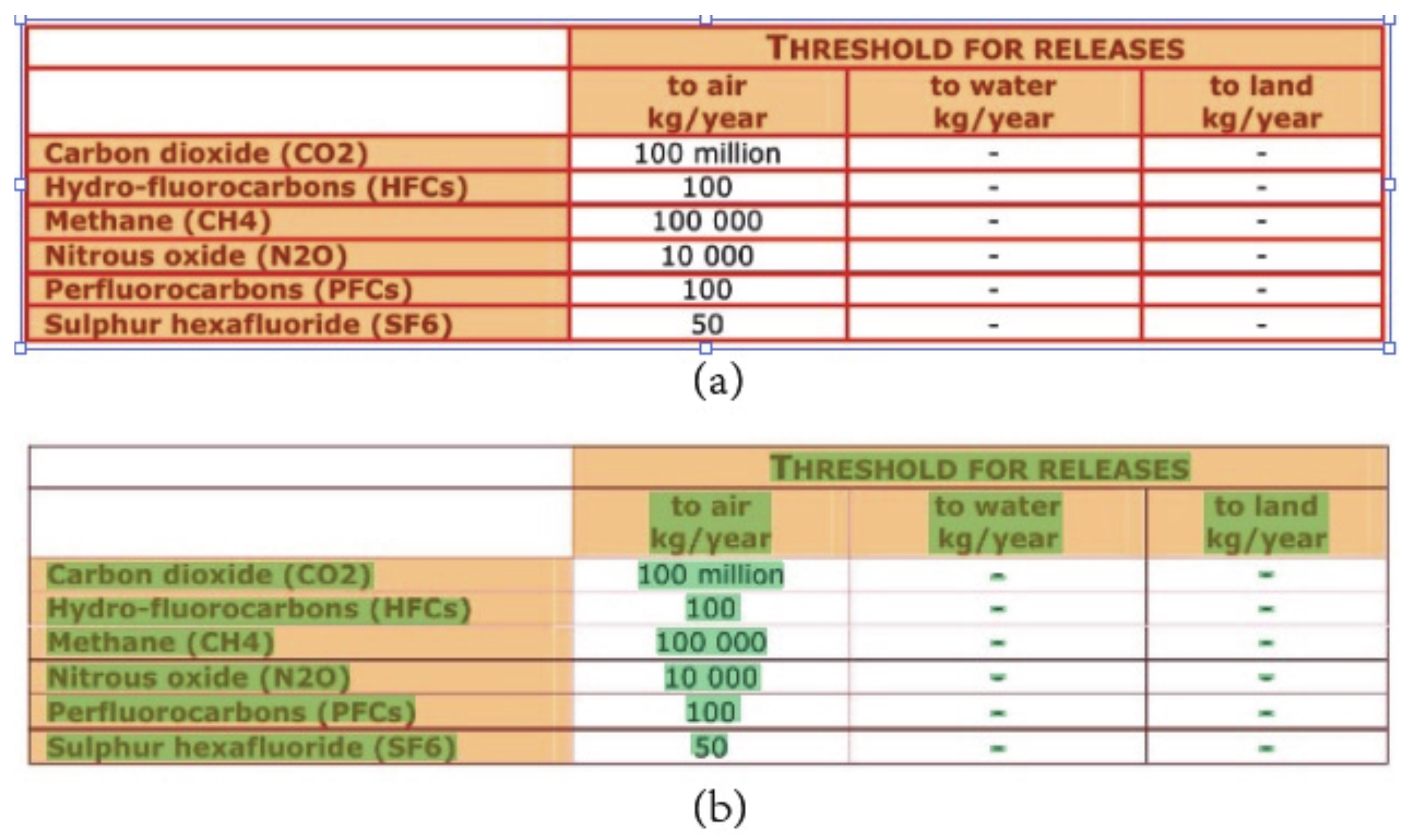}
  \caption{(a): aligned bounding boxes. Red lines are based on the cells' aligned bounding boxes. (b): text-focused bounding boxes. Green areas show cells' bounding boxes.}
  \label{fig:icdar2013_new_annotation}
\end{center}
\end{figure}

To generate text-focused bounding boxes, we use PP-OCR~\cite{du2021pp} to process the aligned bounding boxes further. More specifically, the ground truth of the ICDAR2013 dataset contains the aligned bounding boxes of the tables and the table cells, meaning that we only need to annotate text-focused bounding boxes. PP-OCR is a tool that can detect text areas and recognize the texts in the image. We extract the cell images firstly, then further find the text area in the cropped images to obtain the text-focused bounding boxes. Figure~\ref{fig:icdar2013_new_annotation} shows a sample from ICDAR2013 dataset with two types of bounding boxes. It is worth to note that all of the empty cells are discarded when text-focused bounding boxes are used, and two cells are annotated as "vertically connected" or "horizontally connected" when there are empty cells between them, as discussed earlier.

\subsubsection{The impact of table detection and table cell detection}
In the proposed method, it is assumed that the tables have already been extracted from document images and the bounding boxes of the table cells are known. Indeed, this is a reasonable assumption as there have been many studies achieving promising performance on table detection and table structure detection \cite{jiang2021tabcellnet,FERNANDES2022317}. For many real cases, the input would be a document image, meaning that there is still a need for a table detection model and a table cell detection model as the pre-processing steps of the proposed method. Figure~\ref{fig:demonstration_of_proposed_method} shows a full pipeline of processing a document image which comes from ICDAR2013 dataset. In Figure~\ref{fig:demonstration_of_proposed_method}, Step(1) shows a sample of table detection result; Step(2) extract the table from the document image, feed it into a cell detection model and then obtain the final cells' bounding boxes; Step(3) uses the proposed method to build the associations of table cells and represent the table structure with a graph. To visualize the result of Step(3), in Figure~\ref{fig:demonstration_of_proposed_method}, a query cell is given, and Figure~\ref{fig:demonstration_of_proposed_method} illustrates its corresponding column. Notably, we add extra post-processing methods, such as merging bounding boxes when they have overlap to reduce the errors introduced in Step(2). In this full pipeline, Step(1) almost does not introduce any errors to the final results because many table detection models, such as CascadTabNet~\cite{prasad2020cascadetabnet} report an F1 score of 1.0 under the ICDAR2013 dataset. Meanwhile, in Step(2), the output can be of two types of bounding boxes: 1) aligned bounding box, 2) text-focused bounding box, depending on the design of the table cell detection model. The impact of these two types of bounding boxes is discussed in the previous section. There have been a few studies that achieved promising results, such as TableCellNet~\cite{jiang2021tabcellnet} reporting an F1 score of 0.937, GTE~\cite{Zheng2021GlobalTE} reporting an F1 score of 96.24, but still this step may introduce extra to the table structure recognition result. It is worth mentioning that other bottom-up state-of-the-art methods, such as GraphTSR~\cite{chi2019complicated} have the same assumption with the proposed method in this paper, and the proposed method can outperform these methods by a large margin as mentioned earlier in  Section~\ref{sec:experiment_results}.

\section{Conclusion}
\label{sec:conclusion}
In this paper, we have proposed to use the existing graph-based representation model for complicated table structures and have defined the table structure recognition problem as a cell association classification problem. Our proposed method builds on a Conditional Attention Network (CATT-Net), and it has been shown to outperform the state-of-the-art methods under various public datasets. Furthermore, the proposed CATT-Net method can improve the model performance significantly up to 0.963, 0.981 and 0.982 in terms of Micro-averaged F1 scores on the ICDAR2013, SciTSR, and SciTSR-COMP dataset respectively while those scores of the state-of-the-art method are 0.930, 0.964 and 0.955 on the three datasets. Since the proposed CATT-Net is a bottom-up approach relying on the cells' bounding boxes, whether the type of bounding boxes is aligned or text-focused can significantly influence the model performance. Aligned bounding boxes can improve the structure recognition performance, however they are difficult to obtain by the detection or segmentation models, especially for the tables without borders. We have also discussed the influence of these two types of bounding boxes and further annotated ICDAR2013 dataset providing these two types of bounding boxes simultaneously. 

Tables can provide information through the texts in the cells, and the texts can be also useful features. Therefore, we can extract the texts from document images with the help of OCR tools, and improve the performance using the features from both the image modality and text modality, which can be a direction for the future work.

\bibliographystyle{sn-basic}
\bibliography{sn-bibliography}

\end{document}